\documentclass{article} % For LaTeX2e

\usepackage{iclr2018_conference,times}

\usepackage[T1]{fontenc}
\usepackage[utf8]{inputenc}
\usepackage[english]{babel}  % improved hyphening and internationalization
\usepackage{microtype}       % micro typography!!
\usepackage{textgreek}       % greek symbols in text (i.e. \textalpha}
\usepackage{newunicodechar}  % for defining new unicode characters
\usepackage{xspace}          % provides the \xspace command to fix the space after command problem

\usepackage{subcaption}     % for subfigures
\usepackage{algorithm2e}    % algorithms [linesnumbered]
\usepackage{booktabs}       % professional-quality tables
\usepackage{sidecap}        % for side captions
\usepackage{graphicx}

\sidecaptionvpos{figure}{t} % align side-captions to the top

\usepackage{nicefrac}       % compact symbols for 1/2, etc.
\usepackage{bm}             % boldmath using \bm also for greek symbols
\usepackage{amsmath}        % equations
\usepackage{amssymb}

\usepackage{varioref}
\definecolor{mydarkblue}{rgb}{0,0.08,0.45}
\usepackage{hyperref}
\hypersetup{ %
    breaklinks=true,
    unicode=true,
    pdftitle={},
    pdfauthor={Klaus Greff},
    pdfsubject={},
    pdfkeywords={},
    pdfborder=0 0 0,
    pdfpagemode=UseNone,
    colorlinks=true,
    linkcolor=mydarkblue,
    citecolor=mydarkblue,
    filecolor=mydarkblue,
    urlcolor=mydarkblue,
    pdfview=FitH}
\usepackage[all]{hypcap}
\usepackage{cleveref}
\usepackage{url}             % simple URL typesetting
\addto\extrasenglish{

}

\crefformat{footnote}{#2\footnotemark[#1]#3}

\DeclareMathOperator*{\argmax}{\arg\!\max}

\usepackage{lipsum}  % for fake text
\usepackage{fixme}
\fxsetup{
status=final,
author=,
layout=pdfmargin, % inline, footnote, or pdfnote
theme=color
}

\usepackage[acronym,hyperfirst=false]{glossaries}  % Contrary to common advice that hyperref should be loaded last, glossaries needs to be after hyperref

\newcommand{\createAcronymCmd}[3]{%
\newacronym[user1=#3]{#1}{#1}{#2}
\expandafter\newcommand\csname #1\endcsname{\gls{#1}\xspace}
\expandafter\newcommand\csname #1s\endcsname{\glspl{#1}\xspace}
}

\newacronymstyle{optional-cite}% new style name
{%
\glsgenacfmt
}%
{%

}
\setacronymstyle{optional-cite}
\glsdisablehyper

\createAcronymCmd{RNN}{Recurrent Neural Network}{\citealp{robinson1987utility, werbos1988generalization}}
\createAcronymCmd{CNN}{Convolutional Neural Network}{}
\createAcronymCmd{DCNN}{Decoder CNN?}{}
\createAcronymCmd{NN}{Neural Network}{}
\createAcronymCmd{FIT}{Feature Integration Theory}{}
\createAcronymCmd{RBM}{Restricted Boltzman Machine}{}
\createAcronymCmd{SGD}{Stochastic Gradient Descent}{}
\createAcronymCmd{LSTM}{Long Short-Term Memory}{}
\createAcronymCmd{AMI}{Adjusted Mutual Information}{\citealp{vinh2010information}}
\createAcronymCmd{EM}{Expectation Maximization}{\citealp{dempster1977maximum}}
\createAcronymCmd{RC}{Reconstruction Clustering}{}
\createAcronymCmd{DAE}{Denoising Autoencoder}{\citealp{behnke2001learning, vincent2008extracting}}
\createAcronymCmd{TAG}{iTerative Amortized Grouping}{}
\createAcronymCmd{MLE}{Maximum Likelihood Estimate}{}
\createAcronymCmd{ReLU}{Rectified Linear Unit}{}
\createAcronymCmd{ELU}{Exponential Linear Unit}{}
\createAcronymCmd{GAN}{Generative Adversarial Network}{}
\createAcronymCmd{NPE}{Neural Physcis Engine}{\citealp{chang2016compositional}}
\createAcronymCmd{MDP}{Markov Decision Process}{}
\createAcronymCmd{BPTT}{Back Propagation Through Time}{}

\newacronym{RNEM}{R-NEM}{Relational N-EM}
\newcommand{\RNEM}{\gls{RNEM}\xspace}

\newacronym[user1=\citealp{greff2017neural}]{NEM}{N-EM}{Neural Expectation Maximization}
\newcommand{\NEM}{\gls{NEM}\xspace}
\newacronym[user1=\citealp{greff2017neural}]{RNNEM}{RNN-EM}{RNN Expectation Maximization}

\makenoidxglossaries

\title{Relational Neural Expectation Maximization: Unsupervised Discovery of Objects and their Interactions}

\author{Sjoerd van Steenkiste\\
Swiss AI Lab IDSIA, SUPSI, USI\\
Lugano, Switzerland\\
\texttt{sjoerd@idsia.ch} \\
\And
Michael Chang \thanks{Work performed while at IDSIA.}\\
UC Berkeley \\
Berkeley, United States {\phantom{AAAAA}}\\
\texttt{mbchang@berkeley.edu} \\
\AND
Klaus Greff \\
Swiss AI Lab IDSIA, SUPSI, USI\\
Lugano, Switzerland{\phantom{AAAAAAAAAAAAAAAAAAAAAA}}\\
\texttt{klaus@idsia.ch} \\
\And
J\"urgen Schmidhuber\\
Swiss AI Lab IDSIA, SUPSI, USI\\
Lugano, Switzerland\\
\texttt{juergen@idsia.ch}
}

\iclrfinalcopy % Uncomment for camera-ready version, but NOT for submission.

\begin{document}

\maketitle
\fxnote{We need to fix the author set-up!}
%%%%%%%%%%%%%%%%%%%%%%%%%%%%%%%  Abstract  %%%%%%%%%%%%%%%%%%%%%%%%%%%%%%%%%%%%%%%%%%

\begin{abstract}
Common-sense physical reasoning is an essential ingredient for any intelligent agent operating in the real-world.
For example, it can be used to simulate the environment, or to infer the state of parts of the world that are currently unobserved.
In order to match real-world conditions this causal knowledge must be learned without access to supervised data.
To address this problem we present a novel method that learns to discover objects and model their physical interactions from raw visual images in a purely \emph{unsupervised} fashion.
It incorporates prior knowledge about the compositional nature of human perception to factor interactions between object-pairs and learn efficiently.
On videos of bouncing balls we show the superior modelling capabilities of our method compared to other unsupervised neural approaches that do not incorporate such prior knowledge.
We demonstrate its ability to handle occlusion and show that it can extrapolate learned knowledge to scenes with different numbers of objects.
\end{abstract}

%%%%%%%%%%%%%%%%%%%%%%%%%%%%%%%  Introduction  %%%%%%%%%%%%%%%%%%%%%%%%%%%%%%%%%%%%%%%%%%

\section{Introduction}
Humans rely on common-sense physical reasoning to solve many everyday physics-related tasks~\citep{lake2016building}.
For example, it enables them to foresee the consequences of their actions (simulation), or to infer the state of parts of the world that are currently unobserved. 
This \emph{causal} understanding is an essential ingredient for any intelligent agent that is to operate within the world. 

Common-sense physical reasoning\fxnote{perhaps there is a better insert} is facilitated by the discovery and representation of objects (a \emph{core} domain of human cognition~\citep{spelke2007core})
\fxnote{According to Tom Griffiths, "core knowledge" is controversial: in the rest of the paper, we probably would want to be more careful with the phrase "core knowledge" because "core knowledge" is merely a hypothesis. Munakata1997rethinking gives opposing evidence that babies are inborn with an innate representation of objects. Nevertheless, object-based representations are still a good prior to have anyways, and we should emphasize that. One way Tom suggested we replace ``core knowledge'' is "young infants have surprisingly detailed knowledge of the world around them. Among this knowledge is a notion of intuitive physics."} 
that serve as primitives of a compositional system.
They allow humans to decompose a complex visual scene into distinct parts, describe relations between them and reason about their dynamics as well as the consequences of their interactions~\citep{ullman2017mind,battaglia2013simulation,lake2016building}.

The most successful machine learning approaches to common-sense physical reasoning incorporate such prior knowledge in their design. 
They maintain explicit object representations, which allow for general physical dynamics to be learned between object pairs in a compositional manner~\citep{chang2016compositional, battaglia2016interaction, watters2017visual}. 
However, in these approaches learning is \emph{supervised}, as it relies on object-representations from external sources (e.g. a physics simulator) that are typically unavailable in real-world scenarios.

Neural approaches that learn to directly model motion or physical interactions in pixel space offer an alternative solution~\citep{sutskever2009recurrent, srivastava2015unsupervised}.
However, while unsupervised, these methods suffer from a lack compositionality at the representational level of objects.
This prevents such end-to-end neural approaches from efficiently learning functions that operate on multiple entities and generalize in a human-like way (c.f. \citet{lake2016building, santoro2017simple, battaglia2013simulation}, but see~\citet{perez2017film}).

In this work we propose \RNEM, a novel approach to common-sense physical reasoning that learns physical interactions between objects from raw visual images in a purely \emph{unsupervised} fashion.
At its core is \NEM, a method that allows for the discovery of compositional object-representations, yet is unable to model interactions between objects.
Therefore, we endow \NEM with a relational mechanism inspired by previous work~\citep{santoro2017simple,chang2016compositional,battaglia2016interaction}, enabling it to factor interactions between object-pairs, learn efficiently, and generalize to visual scenes with a varying number of objects without re-training.

%%%%%%%%%%%%%%%%%%%%%%%%%%%%%%%  Model  %%%%%%%%%%%%%%%%%%%%%%%%%%%%%%%%%%%%%%%%%%

\section{Method}
Our goal is to learn common-sense physical reasoning in a purely unsupervised fashion directly from visual observations. 
We have argued that in order to solve this problem we need to exploit the compositional structure of a visual scene.
% Previously (Michael): We have argued that in order to solve this problem we need to exploit its compositional structure. 
Conventional unsupervised representation learning approaches (eg. VAEs~\cite{kingma2013auto}; GANs~\cite{goodfellow2014generative}) learn a single distributed representation that \emph{superimposes} information about the input, without imposing any structure regarding objects or other low-level primitives.
These monolithic representations can not factorize physical interactions between pairs of objects and therefore lack an essential inductive bias to learn these efficiently.
% Michael: a great opportunity to briefly summarize the binding problem here.
Hence, we require an alternative approach that can discover objects representations as primitives of a visual scene in an unsupervised fashion.

One such approach is \emph{Neural Expectation Maximization} (N-EM; \citet{greff2017neural}), which learns a separate distributed representation for each object described in terms of the same features through an iterative process of perceptual grouping and representation learning.
The compositional nature of these representations enable us to formulate \emph{Relational N-EM} (R-NEM): a novel \emph{unsupervised} approach to common-sense physical reasoning that combines \NEM (\autoref{sec:nem}) with an interaction function that models relations between objects efficiently (\autoref{sec:rnem}).

% N-EM
\subsection{Neural Expectation Maximization}
\label{sec:nem}
% High-level summary of the method
\emph{Neural Expectation Maximization} (N-EM;~\cite{greff2017neural}) is a differentiable clustering method that learns a representation of a visual scene composed of primitive object representations.
These representations adhere to many useful properties of a symbolic representation of objects, and can therefore be used as primitives of a compositional system~\citep{hummel2004solution}.
They are described in the same format and each contain only information about the object in the visual scene that they correspond to.
Together, they form a representation of a visual scene composed of objects that is learned in an unsupervised way, which therefore serves as a starting point for our approach. % maybe comprise

% how is its objective achieved: likelihood -> EM -> unrolling
The goal of \NEM is to group pixels in the input that belong to the same object (perceptual grouping) and capture this information efficiently in a distributed representation $\bm{\theta}_{k}$ for each object.
At a high-level, the idea is that if we were to have access to the family of distributions $P(\bm{x}|\bm{\theta}_{k})$ (a statistical model of images given object representations $\bm{\theta}_{k}$) then we can formalize our objective as inference in a mixture of these distributions.
By using \EM to compute a Maximum Likelihood Estimate (MLE) of the parameters of this mixture ($\bm{\theta}_{1}, \hdots, \bm{\theta}_{K}$), we obtain a grouping (clustering) of the pixels to each object (component) and their corresponding representation.
In reality we do not have access to $P(\bm{x}|\bm{\theta}_{k})$, which \NEM learns instead by parameterizing the mixture with a neural network and back-propagating through the iterations of the unrolled generalized \EM procedure. 
% Previously (Michael): In reality we do not have access to $P(\bm{x}|\bm{\theta}_{k})$, which \NEM learns instead by parameterizing the mixture with a neural network, to then back-propagate through the iterations of the unrolled generalized \EM procedure. 

% Likelihood
Following~\cite{greff2017neural}, we model each image $\bm{x} \in \mathbb{R}^{D}$ as a spatial mixture of $K$ components parameterized by vectors $\bm{\theta}_1, \dots, \bm{\theta}_K \in \mathbb{R}^{M}$. 
A neural network $f_{\phi}$ is used to transform these representations $\bm{\theta}_k$ into parameters $\psi_{i,k} = f_{\phi}(\bm{\theta}_k)_i$ for separate pixel-wise distributions.
A set of binary latent variables $\bm{\mathcal{Z}} \in [0, 1]^{D \times K}$ encodes the unknown true pixel assignments, such that $z_{i,k}=1$ iff pixel $i$ was generated by component $k$.
The full likelihood for $\bm{x}$ given $\bm{\theta} = (\bm{\theta}_1, \dots, \bm{\theta}_K)$ is given by:
\begin{equation}
P(\bm{x}|\bm{\theta}) = \prod_{i=1}^{D} \sum_{\bm{z}_i} P(x_i, \bm{z}_i|\bm{\psi}_{i}) = \prod_{i=1}^{D} \sum_{k=1}^K P(z_{i,k}=1) P(x_i|\psi_{i, k}, z_{i,k}=1).
\end{equation}

%%%%%%%%% R-NEM figure %%%%%%
\begin{figure}
\centering
\includegraphics[width=\textwidth]{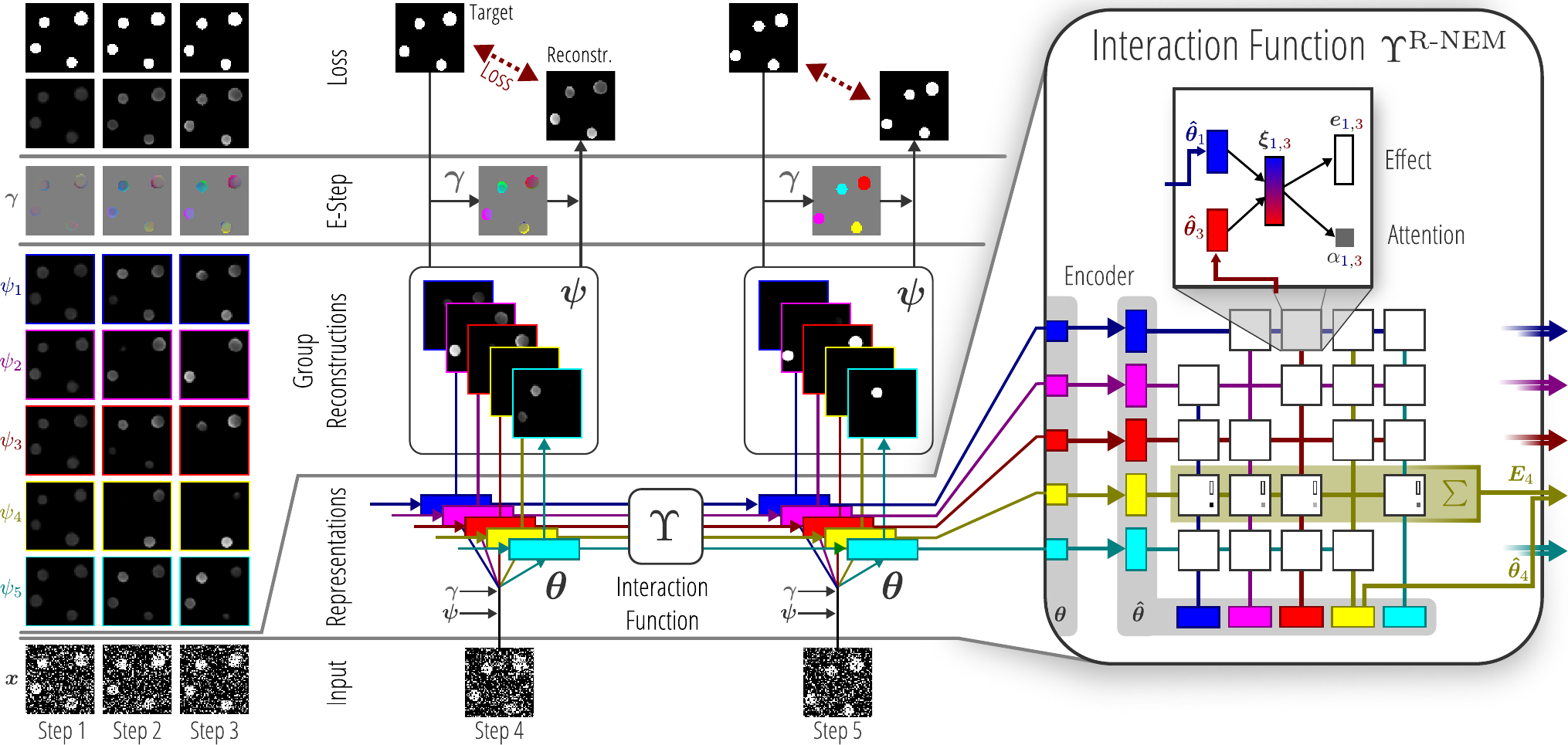}
\caption{Illustration of the different computational aspects of R-NEM when applied to a sequence of images of bouncing balls.
Note that $\bm{\gamma}, \bm{\psi}$ at the \emph{Representations} level correspond to the $\bm{\gamma}$ (\emph{E-step}), $\bm{\psi}$ (\emph{Group Reconstructions}) from the previous time-step. 
Different colors correspond to different cluster components (object representations).% Added by Michael
The right side shows a computational overview of $\Upsilon^{\text{R-NEM}}$, a function that computes the pair-wise interactions between the object representations. 
% Previously (Michael): On the right side, a computational overview of $\Upsilon^{\text{R-NEM}}$ is shown.
} 
\label{fig:r-nem}
\end{figure}
%%%%%%%%%%%%%%%%%%%

% EM
If $f_{\phi}$ has learned a statistical model of images given object representations $\bm{\theta}_{k}$, then we can compute the object representations for a given image $\bm{x}$ by maximizing $P(\bm{x}|\bm{\theta})$.
Marginalization over $\bm{z}$ complicates this process, thus we use generalized \EM to maximize the following lowerbound instead:

\begin{equation}
\mathcal{Q}(\bm{\theta}, \bm{\theta}^{\text{old}}) = \sum_{\mathbf{z}} P(\mathbf{z}|\bm{x}, \bm{\psi}^{\text{old}}) \log P(\bm{x}, \mathbf{z} | \bm{\psi}).
\label{eq:lowerbound}
\end{equation}

Each iteration of generalized EM consists of two steps: the \emph{E-step} computes a new estimate of the posterior probability distribution over the latent variables $\gamma_{i,k} := P(z_{i, k}=1|x_{i},\psi_{i}^{\text{old}})$ given $\bm{\theta}^\text{old}$ from the previous iteration. 
It yields a new soft-assignment of the pixels to the components (clusters), based on how accurately they model $\bm{x}$.
The generalized \emph{M-step} updates $\bm{\theta}^\text{old}$ by taking a gradient ascent step on~\eqref{eq:lowerbound}, using the previously computed soft-assignments: $\bm{\theta}_{k}^{\text{new}} = \bm{\theta}_{k}^{\text{old}} + \eta \cdot \partial\mathcal{Q} / \partial\bm{\theta}_{k}$.\footnote{We can not compute $\argmax_\theta \mathcal{Q}(\bm{\theta}, \bm{\theta}^{\text{old}})$ analytically, due to non-linearity of $f_{\phi}$.}
 
 % unrolling
The unrolled computational graph of the generalized \EM steps is differentiable, which provides a means to train $f_{\phi}$ to implement a statistical model of images given object representations.
Using back-propagation through time (eg.~\citet{werbos1988generalization, williams1989complexity}) we train $f_{\phi}$ to  minimize the following loss:
 
\begin{equation}
L(\bm{x}) = - \sum_{i=1}^D \sum_{k=1}^K
\underbrace{\gamma_{i,k} \log P(x_i, z_{i,k}|\psi_{i,k})}_{\text{intra-cluster loss}} -
\underbrace{(1-\gamma_{i,k}) D_{KL}[P(x_i) || P(x_i|\psi_{i,k}, z_{i,k})]}_{\text{inter-cluster loss}}.
\label{eq:global_loss}
\end{equation} 

The intra-cluster term is identical to~\eqref{eq:lowerbound}, which credits each component for accurately representing pixels that have been assigned to it.
The inter-cluster term ensures that each representation only captures the information about the pixels that have been assigned to it.

A more powerful variant of \NEM can be obtained (RNN-EM) by substituting the generalized M-step with a recurrent neural network having hidden state $\bm{\theta}_{k}$. 
In this case, the entirety of $f_{\phi}$ consists of a recurrent encoder-decoder architecture that receives $\bm{\gamma}_{k}(\bm{x} - \bm{\psi}_{k})$ as input at each step.
% Previously (Michael): In this case the entirety of $f_{\phi}$ consists of a recurrent encoder-decoder architecture that receives $\bm{\gamma}_{k}(\bm{x} - \bm{\psi}_{k})$ as input at each step. 

% learning - Gestalt / object discovery (optional)
The learning objective in~\eqref{eq:global_loss} is prone to trivial solutions in case of overcapacity, which could prevent the network from modelling the statistical regularities in the data that correspond to objects. 
By adding noise to the input image or reducing $\bm{\theta}$ in dimensionality we can guide learning to avert this. 
Moreover, in the case of RNN-EM one can evaluate~\eqref{eq:global_loss} at the following time-step (\emph{predictive coding}) to encourage learning of object representations and their corresponding dynamics.
% Moreover, in case of RNN-EM one can evaluate~\eqref{eq:global_loss} at the following time-step (\emph{predictive coding}) to encourage learning of object representations and their corresponding dynamics.
One intuitive interpretation of using denoising or next-step prediction as part of the training objective is to guide the network to learn about essential properties of objects, in this case those that correspond to the Gestalt Principles of \emph{pr\"{a}gnanz} and \emph{common fate}~\citep{hatfield1985status}.

\subsection{Relational Neural Expectation Maximization}
\label{sec:rnem}

% Short-comings of RNN-EM
RNN-EM (unlike N-EM) is able to capture the dynamics of individual objects through a parametrized recurrent connection that operates on the object representation $\bm{\theta_{k}}$ across consecutive time-steps.
% Previously (Michael): RNN-EM (unlike N-EM) is able to capture the dynamics of objects through a parametrized recurrent connection that operates on the object representation $\bm{\theta_{k}}$ across consecutive time-steps.
However, the relations and interactions that take place \emph{between} objects can not be captured in this way.
In order to overcome this shortcoming we propose \emph{Relational N-EM} (R-NEM), which adds relational structure to the recurrence to model interactions between objects without violating key properties of the learned object representations.

% Our approach high-level (Compositionality needs to be mentioned explicity here, because we have set the stage for that) TODO.
Consider a generalized form of the standard RNN-EM dynamics equation, which computes the object representation $\bm{\theta}_{k}$ at time $t$ as a function of all object representations $\bm{\theta} := [\bm{\theta}_{1}, \hdots, \bm{\theta}_{K}]$ at the previous time-step through an \emph{interaction function} $\Upsilon$:
\begin{equation}
    \bm{\theta}_{k}^{(t)} = \text{RNN}(\tilde{\bm{x}}^{(t)}, \Upsilon_{k}(\bm{\theta}^{(t-1)})) := \sigma(\bm{W} \cdot \tilde{\bm{x}}^{(t)} + \bm{R} \cdot \Upsilon_{k}(\bm{\theta}^{(t-1)})).
\label{eq:recurrent_update}
\end{equation}
Here $\bm{W}, \bm{R}$ are weight matrices, $\sigma$ is the sigmoid activation function, and $\tilde{\bm{x}}^{(t)}$ is the input to the recurrent model at time $t$ (possibly transformed by an encoder). 
When $\Upsilon_{k}^{\text{RNN-EM}}(\bm{\theta}):=\bm{\theta}_{k}$, this dynamics model coincides with a standard RNN update rule, thereby recovering the original RNN-EM formulation.  

% Our approach high-level (Compositionality needs to be mentioned explicitly here, because we have set the stage for that) TODO.
The inductive bias incorporated in $\Upsilon$ reflects the modeling assumptions about the interactions between objects in the environment, and therefore the nature of $\bm{\theta}_k$'s interdependence.\fxnote{klaus: structure as opposed to inductive bias?}
If $\Upsilon$ incorporates the assumption that no interaction takes place between objects, then the $\bm{\theta}_k$'s are fully independent and we recover $\Upsilon^{\text{RNN-EM}}$.
On the other hand, if we do assume that interactions among objects take place, but assume very little about the structure of the interdependence between the $\bm{\theta}_k$'s, then we forfeit useful properties of $\bm{\theta}_k$ such as compositionality.
For example, if $\Upsilon := \text{MLP}(\bm{\theta})$ we can no longer extrapolate learned knowledge to environments with more or fewer than $K$ objects and lose overall data efficiency~\citep{santoro2017simple}.
% Previously (Michael): For example if $\Upsilon := \text{MLP}(\bm{\theta})$ we can no longer extrapolate learned knowledge to environments with more or fewer than $K$ objects and lose overall data efficiency~\citep{santoro2017simple}.
Instead, we can make efficient use of compositionality among the learned object representations $\bm{\theta}_{k}$ to incorporate general but guiding constraints on how these may influence one another~\citep{battaglia2016interaction,chang2016compositional}.
In doing so we constrain $\Upsilon$ to capture interdependence between $\bm{\theta}_{k}$'s in a compositional manner that enables physical dynamics to be learned efficiently, and allow for learned dynamics to be extrapolated to a variable number of objects.

% detail
We propose a parametrized interaction function $\Upsilon^{\text{R-NEM}}$ that incorporates these modeling assumptions and updates $\bm{\theta}_{k}$ based on the pairwise effects of the objects $i \neq k$ on $k$:
\begin{equation}
\begin{split}
& \Upsilon^{\text{R-NEM}}_{k}(\bm{\theta}) = [\bm{\hat{\theta}}_{k};\bm{E}_{k}] \ \ \text{with} \ \ \bm{\hat{\theta}}_{k} = \text{MLP}^\textit{\hspace{1pt}enc}(\bm{\theta}_{k})  \ \ , \ \ \bm{E}_{k} = \sum_{i \neq k} \alpha_{k, i}\cdot \bm{e}_{k,i} \\ &  \ \ \alpha_{k,i} = \text{MLP}^\textit{\hspace{1pt}att}(\bm{\xi}_{k, i}) \ \ , \ \ \bm{e}_{k,i} = \text{MLP}^\textit{\hspace{1pt}eff}(\bm{\xi}_{k, i}) \ \ , \ \ \bm{\xi}_{k, i} = \text{MLP}^\textit{\hspace{1pt}emb}([\bm{\hat{\theta}}_{k};\bm{\hat{\theta}}_{i}])
\end{split}
\label{eq:r-nem}
\end{equation}
where $[\cdot;\cdot]$ is the concatenation operator and $\text{MLP}^{(\cdot)}$ corresponds to a multi-layer perceptron. 
First, each $\bm{\theta}_{i}$ is transformed using $\text{MLP}^\textit{\hspace{1pt}enc}$ to obtain $\bm{\hat{\theta}}_{i}$, which enables information that is relevant for the object dynamics to be made more explicit in the representation.
Next, each pair $(\bm{\hat{\theta}}_{k}, \bm{\hat{\theta}}_{i})$ is concatenated and processed by $\text{MLP}^\textit{\hspace{1pt}emb}$, which computes a shared embedding $\bm{\xi}_{k,i}$ that encodes the interaction between object $k$ and object $i$.
Notice that we opt for a clear separation between the \emph{focus} object $k$ and the \emph{context} object $\emph{i}$ as in previous work~\citep{chang2016compositional}. 
From $\bm{\xi}_{k,i}$ we compute $\bm{e}_{k,i}$: the effect of object $i$ on object $k$; and an attention coefficient $\alpha_{k,i}$ that encodes whether interaction between object $i$ and object $k$ takes place.
These attention coefficients~\citep{bahdanau2014neural,xu2015show} help to select relevant context objects, and can be seen as a more flexible unsupervised replacement of the distance based heuristic that was used in previous work~\citep{chang2016compositional}. % serve as a more flexible unsupervised replacement of the distance based heuristic in~\cite{chang2016compositional} that selects context objects.
Finally, we compute the total effect of $\bm{\theta}_{i\neq k}$ on $\bm{\theta}_{k}$ as a weighted sum of the effects multiplied by their attention coefficient. 
A visual overview of $\Upsilon^{\text{R-NEM}}$ can be seen on the right side of~\autoref{fig:r-nem}.

%%%%%%%%%%%%%%%%%%%%%%%%%%%%%%%  Related Work  %%%%%%%%%%%%%%%%%%%%%%%%%%%%%%%%%%%%%%%%%%

\section{Related Work}
% whole model
Machine learning approaches to common-sense physical reasoning can roughly be divided in two groups: symbolic approaches and approaches that perform state-to-state prediction.
The former group performs inference over the parameters of a symbolic physics engine~\citep{wu2015galileo,ullman2017mind,battaglia2013simulation}, which restricts them to synthetic environments.
The latter group employs machine learning methods to make state-to-state predictions, often describing the state of a system as a set of compact object-descriptions that are either used as an input to the system~\citep{battaglia2016interaction, fragkiadaki2015learning, chang2016compositional,grzeszczuk1998neuroanimator} or for training purposes~\citep{watters2017visual}.
By incorporating information (eg. position, velocity) about objects these methods have achieved excellent generalization and simulation capabilities.
% Although these methods have demonstrated excellent generalization and simulation capabilities they rely on the availability of information (eg. position, velocity) about the objects for their success. 
Purely unsupervised approaches for state-to-state prediction~\citep{sutskever2009recurrent,michalski2014modeling,agrawal2016learning,lerer2016learning} that use raw visual inputs as state-descriptions have yet to rival these capabilities.
% Previous (Michael): Purely unsupervised approaches~\citep{sutskever2009recurrent,michalski2014modeling,agrawal2016learning,lerer2016learning} that use raw visual inputs as state-descriptions have yet to rival these capabilities.
Our method is a purely unsupervised state-to-state prediction method that operates in pixel space, taking a first step towards unsupervised learning of common-sense reasoning in real-world environments.
% Previously (Michael): Our method is a purely unsupervised state-to-state prediction method that operates in pixel space and takes a first step in that direction.

% Interaction Function
The proposed interaction function $\Upsilon^{\text{R-NEM}}$ can be seen as a type of Message Passing Neural Network (MPNN;~\citet{gilmer2017neural}) that incorporates a variant of neighborhood attention~\citep{duan2017one}. In light of other recent work~\citep{zaheer2017deep} it can be seen as a permutation equivariant set function.  

% object / grouping approaches 
R-NEM relies on N-EM~\citep{greff2017neural} to discover a compositional object representation from raw visual inputs.
A closely related approach to N-EM is the TAG framework~\citep{greff2016tagger}, which utilizes a similar mechanism to perform inference over group representations, but in addition performs inference over the group assignments.
In recent work TAG was combined with a recurrent ladder network~\citep{ilin2017recurrent} to obtain a powerful model (RTagger) that can be applied to sequential data. 
However, the lack of a single compact representation that captures all information about a group (object) makes a compositional treatment of physical interactions more difficult.
Other unsupervised approaches rely on attention to group together parts of the visual scene corresponding to objects~\citep{eslami2016attend, gregor2015draw}. These approaches suffer from a similar problem in that their sequential nature prevents a coherent object representation to take shape.
% Michael: an opportunity to talk about the binding problem here

% compositional inspiration
Other related work have also taken steps towards combining the learnability of neural networks with the compositionality of symbolic programs in modeling physics~\citep{chang2016compositional,battaglia2016interaction}, playing games~\citep{kansky2017schema,denil2017programmable}, learning algorithms~\citep{cai2017making,pmlr-v70-bosnjak17a,reed2015neural,li2016neural}, visual understanding~\citep{johnson2017inferring,DBLP:journals/corr/EllisRST17}, and natural language processing~\citep{andreas2016neural,hu2017learning}.
% Other related work has also taken steps towards combining the learnability of neural networks with the compositionality of symbolic programs in modeling physics~\citep{chang2016compositional,battaglia2016interaction}, playing games~\citep{kansky2017schema,denil2017programmable}, learning algorithms~\citep{cai2017making,pmlr-v70-bosnjak17a,reed2015neural,li2016neural}, visual understanding~\citep{johnson2017inferring,DBLP:journals/corr/EllisRST17}, and natural language processing~\citep{andreas2016neural,hu2017learning}.

% sequence plot
\begin{figure}
\centering
\includegraphics[width=\textwidth]{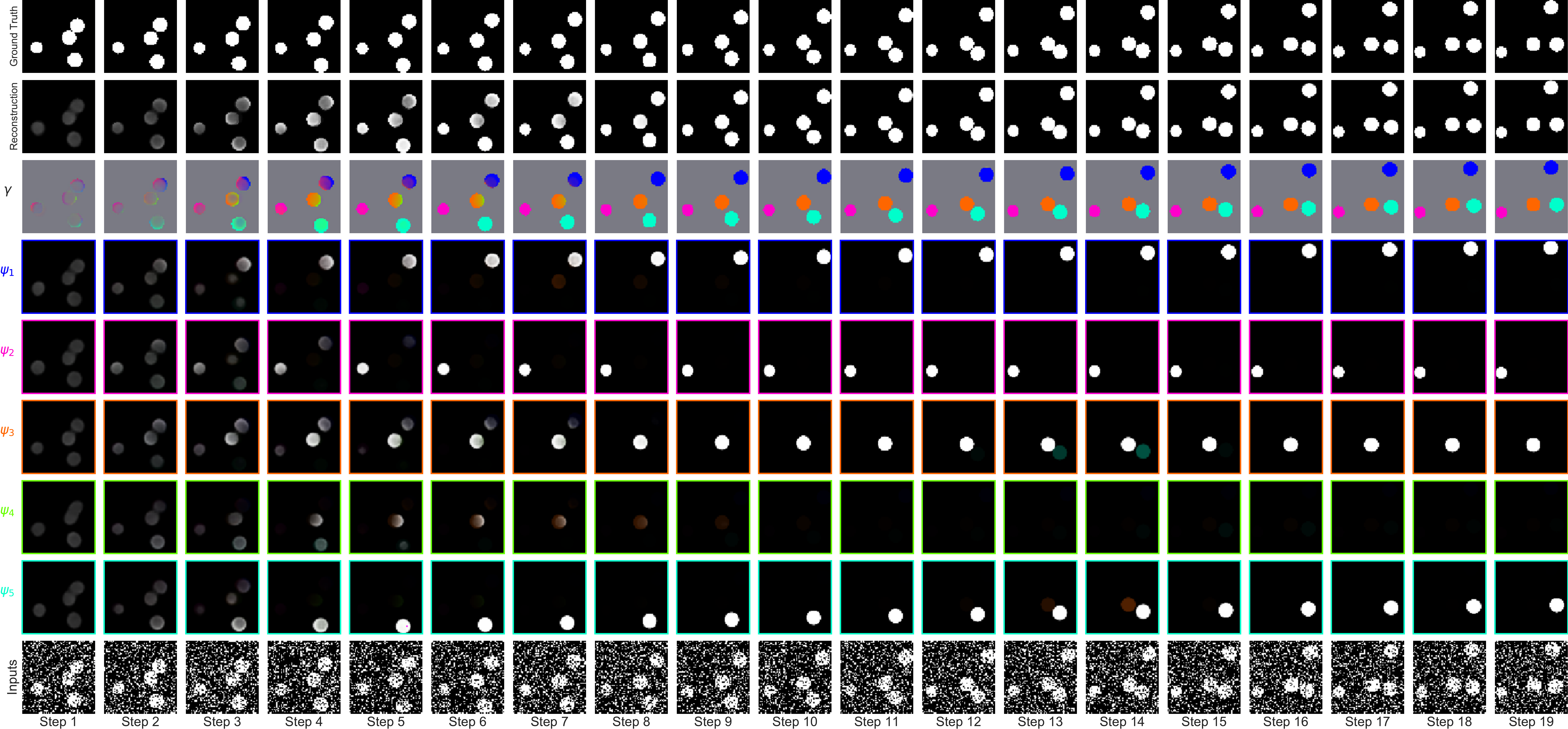}
\caption{R-NEM applied to a sequence of $4$ bouncing balls. Each column corresponds to a time-step, which coincides with an EM step.  
At each time-step, R-NEM computes $K=5$ new representations $\bm{\theta}_{k}$ according to~\eqref{eq:recurrent_update} (see also \emph{Representations} in \autoref{fig:r-nem}) from the input $\bm{x}$ with added noise (bottom row).
From each new $\bm{\theta}_{k}$ a group reconstruction $\bm{\psi}_{k}$ is produced (rows 2-6 from bottom) that predicts the state of the environment at the \emph{next} time-step. Attention coefficients are visualized by overlaying a colored reconstruction of a context object on the white reconstruction of the focus object (see \emph{Attention} in \autoref{paragraph:attention}).
% Previous (Michael): From each new $\bm{\theta}_{k}$ a group reconstruction $\bm{\psi}_{k}$ is produced (rows 2-6) that predicts the state of the environment at the \emph{next} time-step.
Based on the prediction accuracy of $\bm{\psi}$, the \emph{E-step} (see \autoref{fig:r-nem}) computes new soft-assignments $\bm{\gamma}$ (row 7 from bottom), visualized by coloring each pixel $i$ according to their distribution over components $\bm{\gamma}_{i}$.
% Previous (Michael): Based on the prediction accuracy of $\bm{\psi}$, the \emph{E-step} (see \autoref{fig:r-nem}) computes new soft-assignments $\bm{\gamma}$ (row 7), visualized by coloring each pixel $i$ according to their distribution over components $\bm{\gamma}_{i}$.
Row 8 visualizes the total prediction by the network ($\sum_{k}\bm{\psi}_{k} \cdot \bm{\gamma}_{k}$) and row 9 the ground-truth sequence at the next time-step.
}
\label{fig:balls_mass_sequence_plot}
\end{figure}

\section{Experiments}
% reconstruction accuracy plots (BCE , BCE rel, ARI) 
\begin{figure}
\centering
\includegraphics[width=\textwidth]{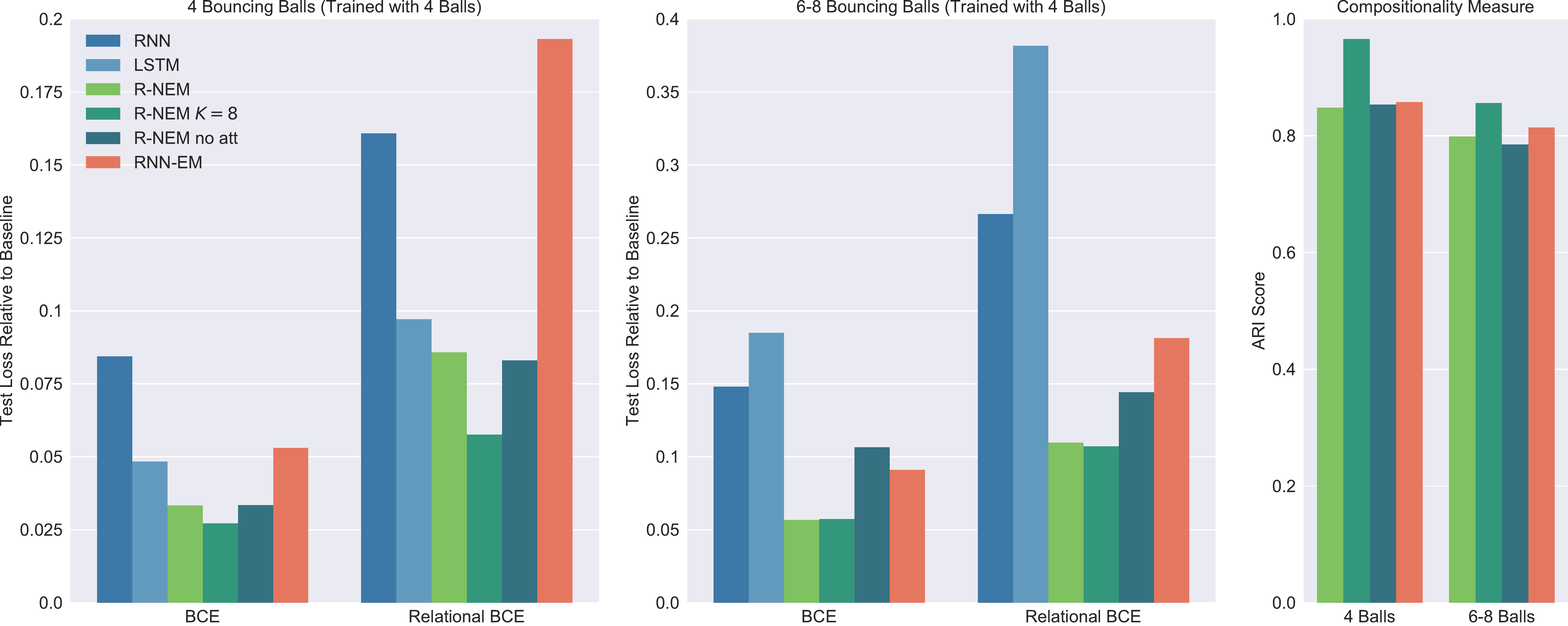}
\caption{Performance of each method on the \emph{bouncing balls} task. Each method was trained on a dataset with 4 balls, evaluated on a test set with $4$ balls (left), and on a test-set with 6-8 balls (middle). The losses are reported relative to the loss of a baseline for each dataset that always predicts the current frame. The ARI score (right) is used to evaluate the degree of compositionality that is achieved.}
\label{fig:bce_plot}
\end{figure}

% Motivation for experiments + overview
In this section we evaluate R-NEM on three different physical reasoning tasks that each vary in their dynamical and visual complexity: bouncing balls with variable mass, bouncing balls with an invisible curtain and the Arcade Learning Environment~\citep{bellemare2013arcade}.
We compare R-NEM to other \emph{unsupervised} neural methods that do not incorporate any inductive biases reflecting real-world dynamics and show that these are indeed beneficial.\footnote{Code is available at \url{https://github.com/sjoerdvansteenkiste/Relational-NEM}.} 

% Global set-up
All experiments use ADAM~\citep{kingma2014adam} with default parameters, on 50K train +  10K validation + 10K test sequences and early stopping with a patience of 10 epochs.
For each of $\text{MLP}^\textit{\hspace{1pt}enc,emb,eff}$ we used a unique single layer neural network with 250 \emph{rectified linear} units.
For $\text{MLP}^\textit{\hspace{1pt}att}$ we used a two-layer neural network: 100 \emph{tanh} units followed by a single \emph{sigmoid} unit.
A detailed overview of the experimental setup can be found in \autoref{app:experiment_details}.
 
% Bouncing Balls Experiment
\paragraph{Bouncing Balls}
% Motivation for experiments + overview
We study the physical reasoning capabilities of R-NEM on the \emph{bouncing balls} task, a standard environment to evaluate physical reasoning capabilities that exhibits low visual complexity and complex non-linear physical dynamics.\footnote{\label{footnote:video}Videos are available at \url{https://sites.google.com/view/r-nem-gifs/}.}
We train R-NEM on sequences of $64 \times 64$ binary images over 30 time-steps that contain four bouncing balls with different masses corresponding to their radii.
The balls are initialized with random initial positions, masses and velocities. 
Balls bounce elastically against each other and the image window.

% explanation seq plot
\paragraph{Qualitative Evaluation}
\autoref{fig:r-nem} presents a qualitative evaluation of R-NEM on the bouncing balls task.
After 10 time-steps it can be observed that the pixels that belong to each of the balls are grouped together and assigned to a unique component (with a saturated color); and that the background (colored grey) has been divided among all components (resulting in a grey coloring).
This indicates that the representation $\bm{\theta}_{k}$ from which each component produces the group reconstruction $\bm{\psi}_{k}$ does indeed only contain information about a unique object, such that together the $\bm{\theta}_{k}$'s yield a compositional object representation of the scene.
The total reconstruction (that combines the group reconstructions and the soft-assignments) displays an accurate reconstruction of the input sequence at the next time-step, indicating that R-NEM has learned to model the dynamics of bouncing balls.

% comparison
\paragraph{Comparison}
% Task desc, losses
We compare the modelling capabilities of R-NEM to an RNN, LSTM~\citep{hochreiter1997long, Gers1999-am} and RNN-EM in terms of the Binomial Cross-Entropy (BCE) loss between the predicted image and the ground-truth image of the last frame,\footnote{Since the E-step in R-NEM and RNN-EM utilizes the ground-truth for reconstruction, we substitute it with a simple $\max$ operator. The resulting loss serves as an \emph{upperbound} to the true BCE loss.} as well as the \emph{relational} BCE that only takes into account objects that currently take part in collision.
Unless specified we use $K=5$.

% reconstruction accuracy test
On a test-set with sequences containing four balls we observe that R-NEM produces markedly lower losses when compared to all other methods (left plot in \autoref{fig:bce_plot}).
Moreover, in order to validate that each component captures only a single ball (and thus compositionality is achieved), we report the Adjusted Rand Index (ARI; \cite{hubert1985comparing}) score between the soft-assignments $\bm{\gamma}$ and the ground-truth assignment of pixels to objects.
In the left column of the ARI plot (right side in \autoref{fig:bce_plot}) we find that R-NEM achieves an ARI score of 0.8, meaning that in roughly $80\%$ of the cases each ball is modeled by a single component.
This suggests that a compositional object representation is achieved for most of the sequences.
Together these observations are in line with our qualitative evaluation and validate that incorporating real world priors is greatly beneficial (comparing to RNN, LSTM) and that $\Upsilon^\text{R-NEM}$ enables interactions to be modelled more accurately compared to RNN-EM in terms of the relational BCE.

Similar to~\cite{greff2017neural} we find that further increasing the number of components during training (leaving additional groups empty) increases the quality of the grouping, see R-NEM $K=8$ in \autoref{fig:bce_plot}.
In addition we observe that the loss (in particular the relational BCE) is reduced further, which matches our hypothesis that compositional object representations are greatly beneficial for modelling physical interactions.

% rollout plot
\begin{figure}
\centering
\includegraphics[width=\textwidth]{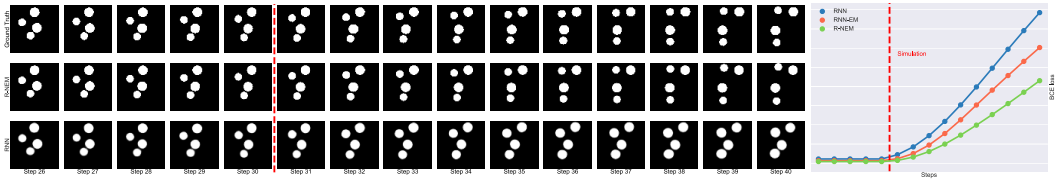}
\caption{\textbf{Left}: Three sequences of 15 time-steps ground-truth (top), R-NEM (middle), RNN (bottom). The last ten time-steps of the sequences produced by R-NEM and RNN are simulated. \textbf{Right}: The BCE loss on the entire test-set for these same time-steps.}
\label{fig:rollout_plot}
\end{figure}

% reconstruction extrapolation
\paragraph{Extrapolating learned knowledge}
We use a test-set with sequences containing 6-8 balls to evaluate the ability of each method to \emph{extrapolate} their learned knowledge about physical interactions between four balls to environments with more balls.
We use $K=8$ when evaluating R-NEM and RNN-EM on this test-set in order to accommodate the increased number of objects.
As can be seen from the middle plot in \autoref{fig:bce_plot}, R-NEM again greatly outperforms all other methods. 
Notice that, since we report the loss relative to a baseline, we roughly factor out the increased complexity of the task. 
Perfect extrapolation of the learned knowledge would therefore amount to \emph{no change} in relative performance.
In contrast, we observe far worse performance for the LSTM (relative to the baseline) when evaluated on this dataset with extra balls. 
It suggests that the gating mechanism of the LSTM has allowed it to learn a sophisticated and overly specialized solution for sequences with four balls that does not generalize to a dataset with 6-8 balls.

R-NEM and RNN-EM scale markedly better to this dataset than LSTM.
Although the RNN similarly suffers to a lesser extend from this type of ``overfitting'', this is most likely due its inability to learn a reasonable solution on sequences of four balls to begin with.
Hence, we conclude that the superior extrapolation capabilities of RNN-EM and R-NEM are inherent to their ability to factor a scene in terms of permutation invariant object representations (see right side of the right plot in \autoref{fig:bce_plot}).

% attention 
\paragraph{Attention} \label{paragraph:attention}
% sequence plot
Further insight in the role of the attention mechanism can be gained by visualizing the attention coefficients, as is done in \autoref{fig:balls_mass_sequence_plot}.
For each component $k$ we draw $\alpha_{k, i} * \bm{\psi}_{i}$ on top of the reconstruction $\bm{\psi}_{k}$, colored according to the color of component $i$.
These correspond to the colored balls (that are for example seen in time-steps 13, 14), which indicate whether component $k$ took information about component $i$ into account when computing the new state (recall~\eqref{eq:r-nem}).
It can be observed that the attention coefficient $\alpha_{k,i}$ becomes non-zero whenever collision takes place, such that a colored ball lights up in the \emph{following} time-steps.
The attention mechanism learned by R-NEM thus assumes the role of the distance-based heuristic in previous work~\citep{chang2016compositional}, matching our own intuitions of how this mechanism would best be utilized.
% MC: maybe we can say more about why this is so cool, if we have space.

% comparison
A quantitative evaluation of the attention mechanism is obtained by comparing R-NEM to a variant of itself that does not incorporate attention (\emph{R-NEM no att}).
\autoref{fig:bce_plot} shows that both methods perform equally well on the regular test set (4 balls), but that \emph{R-NEM no att} performs worse at extrapolating from its learned knowledge (6-8 balls).
A likely reason for this behavior is that the range of the sum in \eqref{eq:r-nem} changes with $K$. %(when trained on a fixed number of balls (4 in our case) the sum in~\eqref{eq:r-nem} operates within a fixed range since it always contains K-1 terms. 
Thus, when extrapolating to an environment with more balls the total sum may exceed previous boundaries and impede learned dynamics.% hence performance deteriorates.
 
\paragraph{Simulation}
Once a scene has been accurately modelled, R-NEM can approximately simulate its dynamics through recursive application of \eqref{eq:recurrent_update} for each $\bm{\theta}_k$.\footnote{Note that in this case the input to the neural network encoder in component $k$ corresponds to $\bm{\gamma}_{k}(\bm{x}^{(t)} - \bm{\psi}^{(t-1)})$, such that the output of the encoder $\tilde{\bm{x}}^{(t)}\approx\bm{0}$ when $\bm{\psi}_{k}^{(t-1)} = \bm{x}^{(t)}$.} 
% This is an example of the same type of \emph{intuitive physical reasoning} that enables people to simulate the environment at a conceptual level in order to predict its state in the immediate future~\citep{lake2016building}.
% to reconstruct a perceptual scene from internal representations~\citep{lake2016building}.
% Although these representations are approximate and incomplete in many ways, they enable humans to simulate the environment at a conceptual level to predict its state in the immediate future.\fxnote{maybe duplicate, see brenden}
% approximate simulation
In \autoref{fig:rollout_plot} we compare the simulation capabilities of R-NEM to RNN-EM and an RNN on the bouncing balls environment.\cref{footnote:video} 
On the left it shows for R-NEM and an RNN a sequence with five normal steps followed by 10 simulation steps, as well as the ground-truth sequence. 
% On the left side of the figure three sequences can be observed, corresponding to the ground-truth sequence (top), the sequence produced by \RNEM (middle), and by an RNN.
% The first five steps rely on external input from the environment, whereas the final ten steps are simulated. % in which case we set the ground-truth equal to the
From the last frame in the sequence it can clearly be observed that \RNEM has managed to accurately simulate the environment. 
Each ball is approximately in the correct place, and the shape of each ball is preserved. 
The balls simulated by the RNN, on the other hand, deviate substantially from their ground-truth position and their size has increased. 
In general we find that \RNEM produces mostly very accurate simulations, whereas the RNN consistently fails. %\footnote{Videos of example simulation sequences are available at \url{www.google.nl}} 
Interestingly we found that the cases in which \RNEM frequently fails are those for which a single component models more than one ball.
The right side of \autoref{fig:rollout_plot} summarizes the BCE loss for these same time-steps across the \emph{entire} test-set.
Although this is a crude measure of simulation performance (since it does not take into account the identity of the balls), we still observe that R-NEM consistently outperforms RNN-EM and an RNN\@. % TODO mention longer timesteps here?
% Michael: can mention that the videos were generated by extrapolating to many more timesteps. I believe that we should mention how many timesteps we can actually rollout to, because the readers would probably care about this.

% Curtain Sequence Plot
\begin{figure}
\centering
\includegraphics[width=\textwidth]
{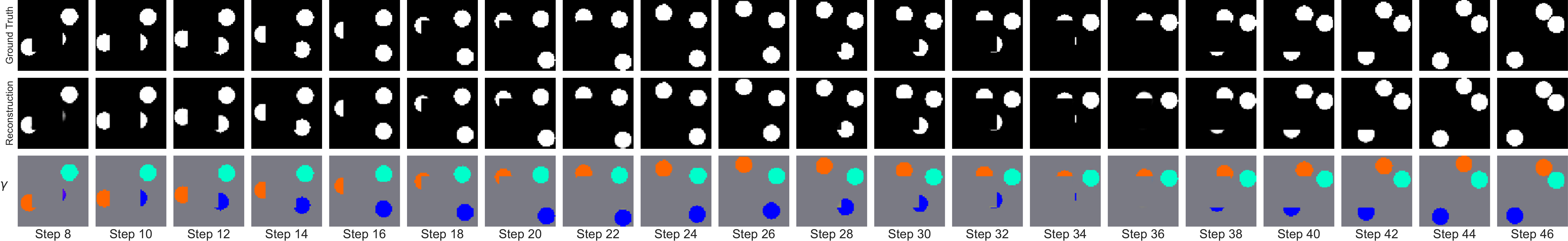}
\caption{R-NEM applied to a sequence of bouncing balls with an invisible curtain. The ground truth sequence is displayed in the top row, followed by the prediction of R-NEM (middle) and the soft-assignments of pixels to components (bottom). R-NEM models objects, as well as its interactions, even when the object is completely occluded (step 36). Only a subset of the steps is shown. }
\label{fig:curtain_sequence_plot}
\end{figure}

% Balls Mass Curtain 
\paragraph{Hidden Factors}
% Explanation
Occlusion is abundant in the real world, and the ability to handle hidden factors is crucial for any physical reasoning system.
We therefore evaluate the capability of \RNEM to handle occlusion using a variant of bouncing balls that contain an invisible ``curtain.''
\autoref{fig:curtain_sequence_plot} shows that \RNEM accurately models the sequence and can maintain object states, even when confronted with occlusion.\cref{footnote:video}
For example, note that in step 36 the ``blue'' ball, is completely occluded and is about to collide with the ``orange'' ball. 
In step 38 the ball is accurately predicted to re-appear at the bottom of the curtain (since collision took place) as opposed to the left side of the curtain.
This demonstrates that \RNEM has a notion of \emph{object permanence} and implies that it understands a scene on a level beyond pixels: it assigns persistence and identity to the objects.

In terms of test-set performance we find that R-NEM (\emph{BCE:} $46.22$, \emph{relational BCE:} $2.33$) outperforms an RNN (\emph{BCE:} $94.64$, \emph{relational BCE:} $4.14$) and an LSTM (\emph{BCE:} $59.32$, \emph{relational BCE:} $2.72$).   

\begin{figure}
\centering
\includegraphics[width=\textwidth]
{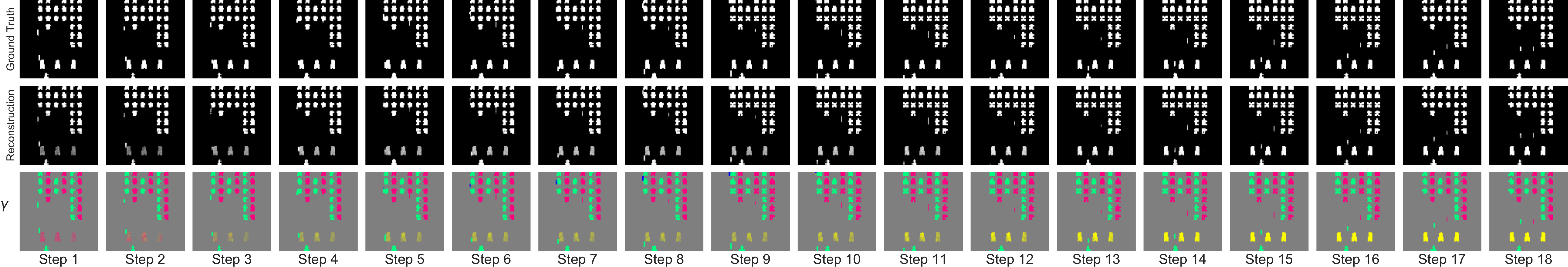}
\caption{R-NEM accurately models a sequence of frames obtained by an agent playing Space Invaders. A group no longer corresponds to an object, but instead assumes the role of high-level entities that engage in similar movement patterns.}
\label{fig:atari_sequence_plot}
\end{figure}

% Atari 
\paragraph{Space Invaders}
To test the performance of R-NEM in a visually more challenging environment, we train it on sequences of $84 \times 84$ binarized images over 25 time-steps of game-play on Space Invaders from the Arcade Learning Environment~\citep{bellemare2013arcade}.\footnote{Binarization ensures that the color group of the entities on the screen does not give away their grouping.}
We use $K=4$ and also feed the action of the agent to the interaction function.
\autoref{fig:atari_sequence_plot} confirms that R-NEM is able to accurately model the environment, even though the visual complexity has increased. 
Notice that these visual scenes comprise a large numbers of (small) primitive objects that behave similarly.
Since we trained R-NEM with four components it is unable to group pixels according to individual objects and is forced to consider a different grouping.
We find that R-NEM assigns different groups to every other column of aliens together with the spaceship, and to the three large ``shields.''
These groupings seem to be based on movement, which to some degree coincides with their semantic roles of the environment.
In other examples (not shown) we also found that R-NEM frequently assigns different groups to every other column of the aliens, and to the three large ``shields.''
Individual bullets and the space ship are less frequently grouped separately, which may have to do with the action-noise of the environment (that controls the movement of the space-ship) and the small size of the bullets at the current resolution that makes them less predictable.

%%%%%%%%%%%%%%%%%%%%%%%%%%%%%%%  Discussion & Conclusion %%%%%%%%%%%%%%%%%%%%%%%%%%%%%%%%%%%%%%%%%%

\section{Discussion and Conclusion}
We have argued that the ability to discover and describe a scene in terms of objects provides an essential ingredient for common-sense physical reasoning.
This is supported by converging evidence from cognitive science and developmental psychology that intuitive physics and reasoning capabilities are built upon the ability to perceive objects and their interactions~\citep{ullman2017mind,spelke1988perceiving}.
% Previously (Michael): This is supported by converging evidence from cognitive science~\citep{ullman2017mind} and developmental psychology that intuitive physics and reasoning capabilities are built upon the ability to perceive objects and their interactions~\citep{spelke1988perceiving}.
The fact that young infants already exhibit this ability, may even suggest an innate bias towards compositionality~\citep{spelke2007core,lake2016building,munakata1997rethinking}.
Inspired by these observations we have proposed R-NEM, a method that incorporates inductive biases about the existence of objects and interactions, implemented by its clustering objective and interaction function respectively.
The specific nature of the objects, and their dynamics and interactions can then be learned efficiently purely from visual observations.

In our experiments we find that R-NEM indeed captures the (physical) dynamics of various environments more accurately than other methods,
and that it exhibits improved generalization to environments with different numbers of objects.
It can be used as an approximate simulator of the environment, and to predict movement and collisions of objects, even when they are completely occluded.
This demonstrates a notion of object permanence and aligns with evidence that young infants seem to infer that occluded objects move in connected paths and continue to maintain object-specific properties~\citep{spelke1990principles}.
Moreover, young infants also appear to expect that objects only interact when they come into contact~\citep{spelke1990principles}, which is analogous to the behaviour of R-NEM to only attend to other objects when a collision is imminent.
In summary, we believe that our method presents an important step towards learning a more human-like model of the world in a completely unsupervised fashion.

% incorporate limitations
Current limitations of our approach revolve around grouping and prediction.
What aspects of a scene humans group together typically varies as a function of the task in mind.
One may perceive a stack of chairs as a whole if the goal is to move them to another room, or as individual chairs if the goal is to count the number of chairs in the stack. 
In order to facilitate this \emph{dynamic} grouping one would need to incorporate top-down feedback from an agent into the grouping procedure to deviate from the built-in inductive biases. 
Another limitation of our approach is the need to incentivize R-NEM to produce useful groupings by injecting noise, or reducing capacity. 
The former may prevent very small regularities in the input from being detected. 
Finally the interaction in the E-step among the groups makes it difficult to increase the number of components above ten without causing harmful training instabilities.
%In general we find that it is more challenging to train R-NEM due to a multitude of interactions and objectives. 
Due to the multitude of interactions and objectives in R-NEM (and RNN-EM) we find that they are sometimes challenging to train. 

In terms of prediction we have implicitly assumed that objects in the environment behave according to rules that can be inferred.
This poses a challenge when objects deform in a manner that is difficult to predict (as is the case for objects in Space Invaders due to downsampling).  
However in practice we find that (once pixels have been grouped together) the masking of the input helps each component in quickly adapting its representation to any unforeseen behaviour across consecutive time steps. 
Perhaps a more severe limitation of R-NEM (and of RNN-EM in general) is that the second loss term of the outer training objective hinders in modelling more complex varying backgrounds, as the background group would have to predict the ``pixel prior'' for every other group. 

We argue that the ability to engage in common-sense physical reasoning benefits any intelligent agent that needs to operate in a physical environment, which provides exciting future research opportunities.
In future work we intend to investigate how top-down feedback from an agent could be incorporated in R-NEM to facilitate dynamic groupings, but also how the compositional representations produced by R-NEM can benefit a reinforcement learner, for example to learn a modular policy that easily generalizes to novel combinations of known objects.
Other interactions between a controller C and a model of the world M (implemented by R-NEM) as posed in~\citet{schmidhuber2015learning} constitute further research directions. 

\section*{Acknowledgements}
The authors wish to thank Tom Griffiths and the anonymous reviewers for helpful comments and constructive feedback.
This research was supported by the Swiss National Science Foundation grant 200021\_165675/1, the EU project ``INPUT'' (H2020-ICT-2015 grant no. 687795), and the Zeno Karl Schindler Foundation Summerschool Grant. Chang would like to thank Christiane Born, Sarah Craver, Cinzia Daldini, and the MIT MISTI Program for supporting his stay in Switzerland.
We are grateful to NVIDIA Corporation for donating us a DGX-1 as part of the Pioneers of AI Research award, and to IBM for donating a ``Minsky'' machine.

%%%%%%%%%%%%%%%%%%%%%%%%%%% Bibliography %%%%%%%%%%%%%%%%%%%%%%%%%%%%%%%%%%%
{\footnotesize
\bibliography{R-NEM}
\bibliographystyle{iclr2018_conference}
}
%%%%%%%%%%%%%%%%%%%%%%%%%%% Appendix %%%%%%%%%%%%%%%%%%%%%%%%%%%%%%%%%%%
\newpage
\appendix

\section{Experiment Details}
\label{app:experiment_details}

In all experiments we train the networks using ADAM~\citep{kingma2014adam} with default parameters, a batch size of 64 and $50\,000$ train +  $10\,000$ validation + $10\,000$ test inputs. 
The quality of the learned groupings is evaluated by computing the Adjusted Rand Index (ARI; \cite{hubert1985comparing}) with respect to the ground truth, while ignoring the background and overlap regions (as is consistent with earlier work~\citep{greff2017neural}). 
We use early stopping when the validation loss has not improved for 10 epochs.

\subsection{Bouncing Balls}
The bouncing balls data is similar to previous work~\citep{sutskever2009recurrent} with a few modifications. 
The data consists of sequences of $64 \times 64$ binary images over 30 time-steps and balls are randomly sampled from two types: one ball is six times heavier and 1.25 times larger in radius than the other. 
The balls are initialized with random initial positions and velocities. 
Balls bounce elastically against each other and the image window. 

% Each input consists of a sequence of binary $64 \times 64$ images containing a fixed number of balls that start in random positions and move along randomly sampled trajectories within the image for 30 steps. \fxnote{TODO Michael: details about the simulator here, e.g. radomly sampled mass / radii}

% network
As in previous work~\citep{greff2017neural} we use a convolutional encoder-decoder architecture with a recurrent neural network as bottleneck, that is updated according to~\eqref{eq:recurrent_update}:
\begin{enumerate}
\item $4 \times 4$ conv. 16 ELU. stride 2. layer norm
\item $4 \times 4$ conv. 32 ELU. stride 2. layer norm
\item $4 \times 4$ conv. 64 ELU. stride 2. layer norm
\item fully connected. 512 ELU. layer norm
\item recurrent. 250 Sigmoid. layer norm on the output
\item fully connected. 512 RELU. layer norm
\item fully connected. $8\times8\times64$ RELU. layer norm
\item $4 \times 4$ reshape 2 nearest-neighbour, conv. 32 RELU. layer norm  
\item $4 \times 4$ reshape 2 nearest-neighbour, conv. 16 RELU. layer norm  
\item $4 \times 4$ reshape 2 nearest-neighbour, conv. 1 Sigmoid  
\end{enumerate}

Instead of using transposed convolutions (to implement the "de-convolution") we first reshape the image using the default nearest-neighbour interpolation followed by a normal convolution in order to avoid frequency artifacts~\citep{odena2016deconvolution}. Note that we do not add layer norm on the recurrent connection. 

At each timestep $t$ we feed $\gamma_{:, k}(\bm{\psi}_{:, k}^{(t-1)} - \bm{\hat{x}}^{(t)})$ as input to the network, where $\bm{\tilde{x}}$ is the input with added bitflip noise ($p=0.2$).
Consistent with earlier work~\citep{greff2017neural} R-NEM is trained with a next-step prediction objective, the prior for each pixel in the data is set to a Bernoulli distribution with $p=0$, and we prevent conflicting gradient updates by not back-propagating any gradients through $\gamma$.  

The Interaction Function $\bm{\Upsilon}^{\text{R-NEM}}$network is structured as follows:

\begin{itemize}
\item $\text{MLP}^{\hspace{1pt}\textit{enc}}$: fully connected. 250 RELU. layer norm
\item $\text{MLP}^{\hspace{1pt}\textit{emb}}$: fully connected. 250 RELU. layer norm
\item $\text{MLP}^{\hspace{1pt}\textit{eff}}$: fully connected. 250 RELU. layer norm
\item $\text{MLP}^{\hspace{1pt}\textit{att}}$: fully connected. 100 Tanh. layer norm - fully connected. 1 Sigmoid.
\end{itemize}

We experimented with deeper architectures, but were unable to observe significant improvement. 

\paragraph{Comparison and Extrapolation}
In the comparison experiment both R-NEM and RNN-EM are trained with $K=5$ (unless otherwise mentioned), following insights from~\cite{greff2017neural}.
On the extrapolation task we adjusted the number of components at test time to $K=8$.

When comparing to RNN-EM we used $\bm{\Upsilon} = \bm{\Upsilon}^{\text{RNN-EM}}$.
For comparing to RNN we set $K=1$, and used $\bm{\Upsilon} = \bm{\Upsilon}^{\text{RNN-EM}}$, yielding a standard recurrent autoencoder that receives at each time-step the difference between the prediction and the noisy ground-truth as input. 
In case of LSTM, we additionally replace the recurrent layer with an LSTM update.
The R-NEM \emph{no att} model is the same as R-NEM, without $\text{MLP}^{\hspace{1pt}\textit{att}}$, such that $\alpha_{:,:} = 1$

\paragraph{Simulation}
Since the E-step relies on the ground-truth, which was not available for simulation, we used a thresholded version of $\max_{k} \bm{\psi}$ at 0.1 (such that everything below becomes 0 and everything above becomes 1) as a replacement in stead.

\paragraph{Occlusion}
On the occlusion dataset we used three balls with equal mass. 
The curtain was spawned at a random location for each sequence. 
We trained R-NEM with $K=5$. 

\subsection{Space Invaders}

We used a pre-trained DQN to produce a dataset with sequences of 25 time-steps. 
The DQN receives a stack of four frames as input and we recorded every first frame of this stack.
These frames were first pre-processed as in~\cite{mnih2013playing} and then thresholded at $0.0001$ to obtain binary images.

Since the images are $84 \times 84$ we used a different encoder and decoder, given by:

\begin{enumerate}
\item $4 \times 4$ conv. 16 ELU. stride 2. layer norm
\item $4 \times 4$ conv. 32 ELU. stride 2. layer norm
\item $4 \times 4$ conv. 32 ELU. stride 2. layer norm
\item $4 \times 4$ conv. 32 ELU. stride 2. layer norm
\item fully connected. 512 ELU. layer norm
\item recurrent. 250 Sigmoid. layer norm on the output
\item fully connected. 512 RELU. layer norm
\item fully connected. $8\times8\times64$ RELU. layer norm
\item $4 \times 4$ reshape 2 nearest-neighbour, conv. 32 RELU. layer norm  
\item $4 \times 4$ reshape 2 nearest-neighbour, conv. 32 RELU. layer norm  
\item $4 \times 4$ reshape 2 nearest-neighbour, conv. 16 RELU. layer norm  
\item $4 \times 4$ reshape 2 nearest-neighbour, conv. 1 Sigmoid  
\end{enumerate}

We used the same architecture for $\Upsilon^{\text{R-NEM}}$, with the only difference that at each time-step we concatenated an embedding of the action produced by the agent to the hidden state. 
Here we used a single layer MLP with 10 units and a \emph{ReLU} activation function to compute this embedding. 

In the Atari experiment we trained with $K=4$ and reduced the input noise to 0.02, in order to preserve tiny elements such as bullets (that only occupy 1-2 pixels).

\end{document}